# Assessing Semantic Annotation Activities with Formal Concept Analysis

Juan Cigarrán-Recuero, Joaquín Gayoso-Cabada, Miguel Rodríguez-Artacho, María-Dolores Romero-López, Antonio Sarasa-Cabezuelo, José-Luis Sierra

*Abstract*—This paper describes an approach to assessing semantic annotation activities based on formal concept analysis (FCA). In this approach, annotators use taxonomical ontologies created by domain experts to annotate digital resources. Then, using FCA, domain experts are provided with concept lattices that graphically display how their ontologies were used during the semantic annotation process. In consequence, they can advise annotators on how to better use the ontologies, as well as how to refine them to better suit the needs of the semantic annotators. To illustrate the approach, we describe its implementation in @note, a Rich Internet Application (RIA) for the collaborative annotation of digitized literary texts, we exemplify its use with a case study, and we provide some evaluation results using the method.

*Keywords*-Semantic Annotation, Formal Concept Analysis, Ontology, Annotation Tool

## 1 Introduction

The enormous efforts to digitize physical resources (documents, books, museum exhibits, etc.), along with recent advances in information and communication technologies, have democratized access to a cultural, scientific and academic heritage previously available to only a few. Likewise, the current trend is to produce new resources in a digital format (e.g., in the context of social networks), which entails an in-depth paradigm shift in almost all the humanistic, social, scientific and technological fields. In particular, the field of the humanities is one which is going through a significant transformation as a result of these digitalization efforts and the paradigm shift associated with the digital age. Indeed, we are witnessing the emergence of a whole host of disciplines, those of Digital Humanities (Berry 2012), which are closely dependent on the production and proper organization of digital collections.

As a result of the undoubted importance of digital collections in modern society, the search for effective and efficient methods to carry out the production, preservation and enhancement of such digital collections has become a key challenge in modern society (Calhoun, 2013). In particular, the annotation of resources with metadata that enables their proper cataloging, search, retrieval and use in different application scenarios is one of the key elements to ensuring the profitability of these collections of digital objects. While the cataloging and retrieval of resources (whether digital or non-digital) have been the object of study in library sciences for decades (Calhoun, 2013), modern applications require annotating resources in semantically richer and more flexible ways, in many cases allowing multiple alternative annotations in the same collection. In consequence, the tendency is to introduce the use of ontology-based semantic technologies, in addition to conventional metadata schemas (Keyser, 2012).

While in recent years we have witnessed significant advances in the automatic annotation of resources, in particular of those with heavy text content (see section 6), there are multiple scenarios in which resource annotation cannot be inferred from the contents of these resources (e.g., scenarios involving resources in which the content is not directly related to the meta-information required). In these cases it is necessary to involve human annotators in the semantic annotation of the resources. The resulting activities are referred to as *semantic annotation activities* in this paper. Some examples of semantic annotation activities are the annotation of digital educational resources (e.g., *learning objects*) in the eLearning domain (Aroyo & Dicheva. 2004, Devedzic et al. 2007, Tiropanis et al. 2009, Kurilovas et al. 2014), the annotation of media content in the multimedia domain (Labra et al. 2010, Mu 2010, Hunter & Herber, 2010, Šimko et al. 2013), or the one chosen as a case study in this paper: the annotation of digitized literary texts (Azouaou & Desmoulins, 2006; Koivunen 2005; Rocha et al. 2009; Schroeter et al. 2006; Tazi et al. 2003; Gayoso et al. 2012, 2013; Donato et al. 2013).

The main objective of any semantic annotation activity should be to produce an annotation of the resources in the underlying digital collection that satisfies all the requirements of accuracy, completeness and adequacy posed by the intended uses of the collection. Therefore, being able to assess to what extent these requirements are accomplished is an obligation in order to guarantee the quality of the final annotation outcomes. On one hand, the result of this assessment could help annotators to make a better use of the semantic models (i.e., the *annotation ontologies*) during the annotation of the resources. On the other hand, it could also be useful to the creators of the ontologies (i.e., the experts in the domain), who could identify how their ontologies should be modified, augmented or refined on the basis of the actual use of these assets during the annotation process. However, for huge collections or dense and semantically-rich annotations, the accomplishment of this assessment by individual inspection of every single annotated resource can become a titanic task. Therefore, providing automatic or semi-automatic assistance in the assessment of semantic annotation activities is an overriding concern in guaranteeing the quality of the annotations performed.

This paper addresses the formulation of mechanisms that support the assessment of semantic annotation activities, in order to enable: (i) better guidance of annotators during the annotation process, and (ii) the iterative refinement of the annotation ontologies. For this purpose, it presents a method of assessing the use of ontologies in semantic annotation activities, based on formal concept analysis (FCA). In this approach, annotators are provided with ontologies specifically designed by domain experts, and they use these ontologies to annotate a collection of digital resources. Then, the annotated collections are automatically analyzed using FCA to allow domain experts access to a lattice-based graphical representation that summarizes the overall annotation activity. This representation is linked to the concepts in the ontology so that at a glance, domain experts can assess how the proposed ontology is being used by annotators. Along with other aspects, they can see which concepts are not being used, which concepts are always used together, and which concepts are used more often than others. As a result, they can provide guidance to the annotators, enabling them to better use the ontologies proposed, or they can find aspects of the ontology that can be improved (e.g., several concepts might be combined into a single concept or they could include new concepts made apparent from the concept lattice). Therefore, and under reasonable assumptions, FCA provides domain experts with the machinery necessary to address the assessment of semantic annotation activities, at least to a semi-automatic extent.

The approach proposed in this paper has been successfully used in @note, a Rich Internet Application (RIA) for the collaborative annotation of digitized literary texts for educational purposes. In @note, teams of annotators (students, in this case) must complete the annotation of digitized literary works with free-text notes, and they must catalogue these notes using concepts taken from an ontology provided by the-domain experts (teachers, in this case). Once the annotation activity is complete, and according to the aforementioned approach, @note allows teachers to examine how students performed the annotation activity by showing them a concept lattice created by considering notes as objects and ontology concepts as attributes in a formal context.

The remainder of this paper is organized as follows. In section 2, we describe the annotation assessment approach. In section 3, we describe its implementation in @note. In section 4, we present a case study, i.e., an annotation activity of a literary work (*The Library of Babel*, a short story authored by the Argentinian writer Jorge Luis Borges). In section 5, we present some evaluation results. In section 6, we describe some related works. Finally, in section 7, we present the conclusions and directions for future work.

## 2 The Assessment Approach

This section describes our approach to the assessment of semantic annotation activities using FCA. In subsection 2.1, we summarize the elements of FCA required in the approach. In subsection 2.2, we present an overview of such an approach. In subsection 2.3, we describe the nature of annotation ontologies. Finally, in subsection 2.4, we present the use of FCA to facilitate the assessment of annotation activities by domain experts.

### 2.1 The elements of FCA

The annotation assessment approach proposed in this paper relies heavily on the construction of concept lattices from annotated digital resources. As mentioned earlier, we use the well-known FCA technique. FCA is a mathematical theory of concept formation derived from lattice and ordered set theories that provides a theoretical model for organizing information and revealing relationships (Wille, 1982) (Wille, 1992) (Ganter, 1999). The main construct of the theory is the *formal concept,* which is derived from a *formal context*.

A *formal context* can be defined as a set of objects, a set of attributes and a set of *is-a* or *has-a* relationships between objects and attributes. A formal concept is a pair $(A, B)$, where $A$ is a set of objects (also known as the *extent* of the formal concept), and $B$ is a set of attributes (also known as the *intent* of the formal concept). The extent and the intent of a formal concept are connected as follows:

- The extent $A$ consists of all the objects that are related to all the attributes in the intent $B$.

- The intent $B$ consists of all the attributes shared by the objects in the extent $A$.

Formal concepts can be ordered by their extents. More formally, $(A, B) \subseteq (C, D) \Leftrightarrow A \subseteq C$; in this case, (C, D) is called a *super-concept* of $(A, B)$ and, conversely, $(A, B)$ a *sub-concept* of (C, D). This ordered relationship is a generalization-specialization, and it can be proven to be a *lattice* (i.e., a concept lattice) based on the basic theorem of FCA (Wille, 1992) (Ganter, 1999).

In a concept lattice, two important types of formal concepts are *object concepts* and *attribute concepts*:

- The *object concept* associated with an object $o$ is the most specific concept that includes $o$ in its extent. The intent of an object concept is defined by all the attributes of $o$, whereas the extent contains not only object $o$ but also all those objects related to all the attributes of $o$.

- The *attribute concept* associated with attribute $a$ is the most generic concept that includes $a$ in its intent. Its extent contains all the objects with attribute $a$, and its intent is defined by all the attributes shared by the objects belonging to the extent set.

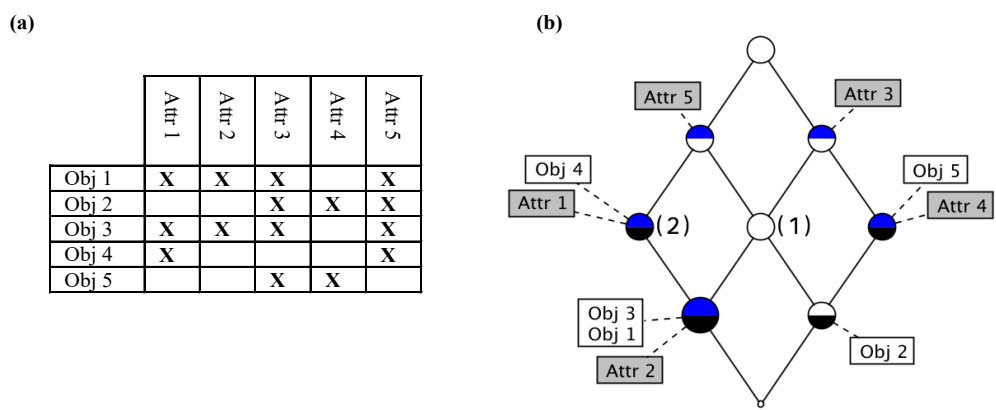

**(a)**

| | Attr 1 | Attr 2 | Attr 3 | Attr 4 | Attr 5 |
|---|---|---|---|---|---|
| Obj 1 | X | X | X | | X |
| Obj 2 | | | X | X | X |
| Obj 3 | X | X | X | | X |
| Obj 4 | X | | | | X |
| Obj 5 | | | X | X | |

Figure 1. (a) A sample formal context and (b) the concept lattice associated with the formal context in (a)

Because concept lattices are ordered sets, they can be displayed naturally in terms of *Hasse diagrams* (Ganter, 1999). In a Hasse diagram: a) there is exactly one node for each formal concept; b) if, for concepts *C1* and *C2*, *C1* $\subseteq$ *C2* holds, then *C2* is placed above *C1*; and c) if *C1* $\subseteq$ *C2* but there is no other concept *C3* such that *C1* $\subseteq$ *C3* $\subseteq$ *C2*, there is a line joining *C1* and *C2*.

Figure 1a shows an example of a formal context, and Figure 1b shows its associated concept lattice using a Hasse diagram[1]. This example illustrates that Hasse diagrams are particularly useful for visualizing concept lattices; thus they will be used in our approach as the primary means of presenting lattices to domain experts. The example also illustrates the method of marking the diagrams to facilitate the identification of formal concepts. To avoid an overloaded representation, each formal concept (i.e., a node in the diagram) is depicted with a minimal set of objects and a minimal set of attributes. Hence, each formal concept can be easily reconstructed from the diagram as follows:

- The extent is given by the union of all the objects depicted in the nodes on the paths leading from the formal target concept to the bottom concept in the diagram. For example, the extent of the formal concept associated with the node marked (1) in Figure 1b is {*obj 1*, *obj 2*, *obj 3*}.

- The intent is given by all the attributes depicted by the nodes on the paths from the formal target concept to the top node in the diagram. For example, in Figure 1b, the intent of the concept represented in node (1) is {*attr 3*, *attr 5*}.

---

[1] Concept lattices in section 2 have been generated with the ConExp application (http://conexp.sourceforge.net/).

In Figure 1b, concept (2) is an object concept of *obj 4* defined as ({*obj 1*, *obj 3*, *obj 4*}, {*attr 1*, *attr 5*}) (object concepts in Figure 1b are represented by coloring the lower half of the node), which means this concept is the most specific concept containing *obj 4* in its extent. Concept (2) is also the attribute concept of *attr 1* (attribute concepts are represented in Figure 1b by coloring the upper half). Concept (1), described as ({*obj 1*, *obj 2*, *obj 3*}, {*attr 3*, *attr 5*}), is neither an object nor an attribute concept.

Quantitative information can also be attached to each node (e.g., the absolute size of the extent of each concept, or its percentage with respect to the overall number of objects).

Figure                                                                                                       2.

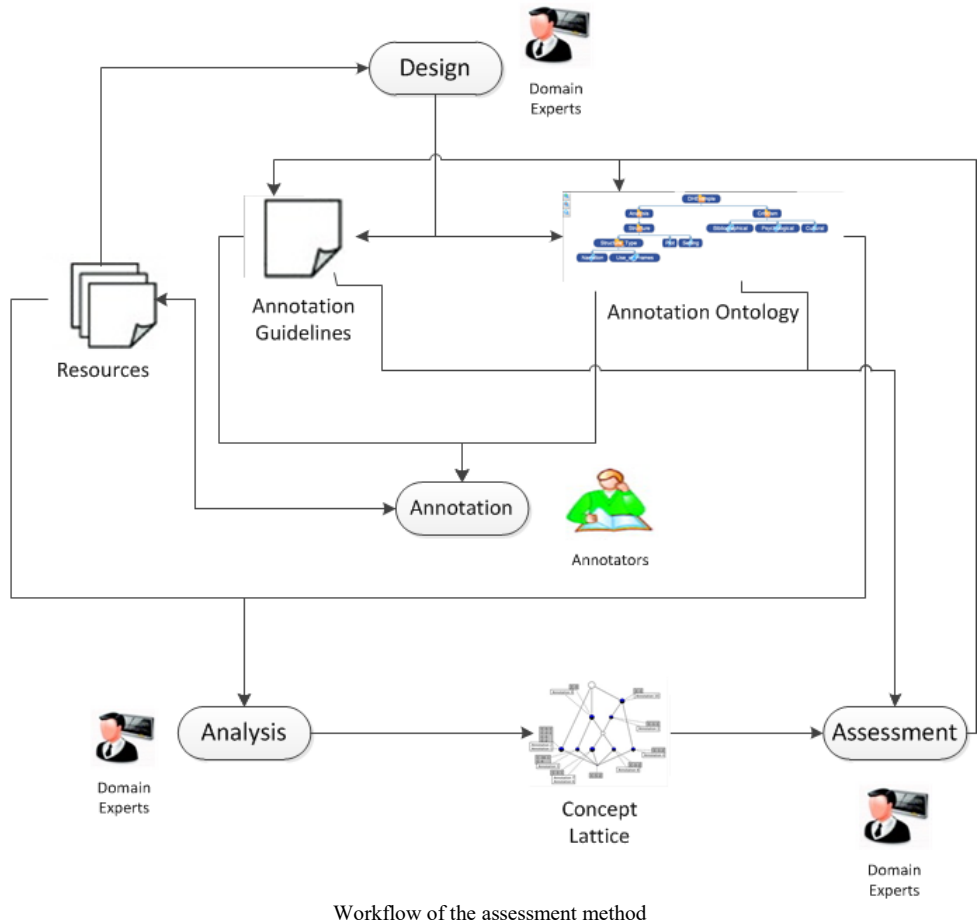

Workflow of the assessment method

### 2.2    Overview of the approach

Figure 2 outlines our approach to the assessment of annotation activities with FCA. The steps in the approach are as follows:

- The *domain experts design* the annotation activities. They begin by analyzing the collections of digital *resources* to be annotated and providing suitable formal *annotation ontologies* for them. They also create suitable *annotation guidelines* to be followed while performing the annotations.

- The *annotators* use the formal ontologies designed by the domain experts to perform the *annotation* of digital resources. During this process, they tag resources with one of several concepts selected from the ontology.

- Once the annotation process is complete, the resources annotated can be automatically analyzed by using FCA (Wille, 1992) (Ganter, 1999). In the *analysis*, the digital resources are the objects of a formal context, and the ontology concepts used to tag them are the attributes of the context. Thus, a *concept lattice* associated with the annotated collection can be automatically constructed.

- The concept lattice is then graphically inspected by the domain experts to assess how their ontologies were used for annotation. As result of this *assessment* stage, domain experts can contribute to improvements in the annotation process. On the one hand, they can advise annotators on the better use of the ontologies by refining the annotation guidelines. On the other hand, they can re-structure their ontologies to better accommodate the pragmatic needs of the annotators. Ontology re-structuring can include the elimination of unused concepts, the fusion of concepts that are commonly used together, the addition of new concepts revealed by the concept lattice, etc.

In addition, driven by the concept-lattice supported assessment, the process can be applied several times. The result is an ontology better suited to the real-world needs of digital-resource annotators.

### 2.3 Annotation Ontologies

Our approach promotes the iterative provision of annotation ontologies. For this purpose:

- During the design step, domain experts provide an initial version of the annotation ontology. To this end, experts begin by characterizing the annotation activity itself. This annotation activity is characterized in terms of: (i) the digital resources to be annotated, (ii) the agents that must carry out the annotation process (i.e., the annotators), and (iii) the goals of the annotation (these goals can vary, depending on the application; typically, annotated resources enable semantic searching and retrieval and ontology-driven semantic browsing) Then, following a conventional, conceptualization-oriented, ontology design process, they provide an ontology specifically oriented to the features of the annotation task.

- The initial ontology is refined as a consequence of the *assessment* stage. As we will indicate later in the paper, this refinement is basically structural: adding new concepts associated to the combination of existing ones, removing useless concepts, melding equivalent concepts into a single one, etc. It lets domain experts solve structural design misconceptions and mistakes on the basis of the evidence gathered from the use of the ontology.

(a)                                                                                    (b)

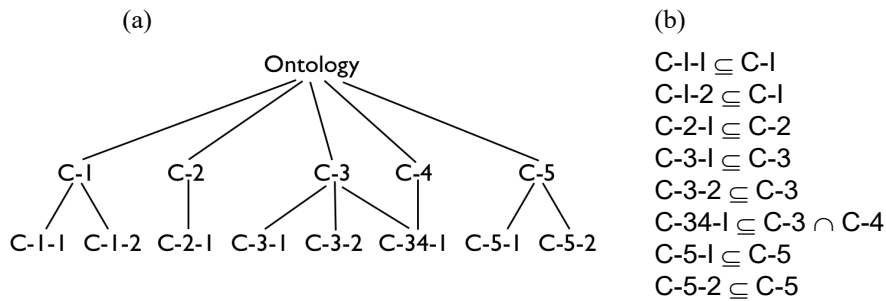

Figure 3. (a) A simple annotation ontology; (b) description logic formalization of (a)

While our proposal is iterative in nature, both the initial provision of the ontology and its refinement after each assessment step require the basic conceptualization skills of domain experts. To make this assumption feasible, it is not reasonable to presuppose domain experts with advanced computer science knowledge, knowledge representation education, formal logic knowledge or experience in artificial intelligence, but educated professionals with profound knowledge of the requirements of a particular annotation task. Therefore, we constrain the shape of the ontologies to single concepts arranged in a multiple inheritance hierarchy. Figure 3a shows an (abstract) example of this kind of ontology, while Figure 3b details its description logic formalization (Baader et al. 2010). As this example makes apparent, although multiple inheritance is allowed (in the example, there is an ontology concept -i.e., C-34-1- that has more than one parent concept, i.e., C-4 and C-3), other kinds of relationships between concepts beyond the *is-a* relationship are intentionally avoided in order to facilitate the active engagement of domain experts in the design step. While it may hinder capturing more complex conceptualizations, hierarchies of this type constitute, on one hand, the skeleton of more sophisticated ontologies (Breuster et al. 2004), and, on the other hand, are sufficiently simple to be authored by domain experts with easy-to-use hierarchy editors (Noy et al., 2001), who can determine the concepts required to organize the resources, as well as to arrange these concepts in meaningful taxonomies. (We also found this to be true in our experience with @note, where the gap between domain experts and computer science knowledge was especially noticeable.)

Finally, it is worthwhile to analyze the potential ontological disagreements among different domain experts in the context of our proposal. On one hand, when several domain experts work together on the definition of an annotation activity, our approach enforces the need to reach consensus before passing on to the annotation step. For this purpose, domain experts can take advantage of suitable collaboration mechanisms such as those available, for instance, in @note. On the other hand, although a particular annotation activity requires a single, consensual annotation ontology, it is important to point out that different domain experts can define different annotation ontologies for the annotation of the *same* body of digital resources. The point here is that each (possibly discordant) ontology is oriented to a *different* annotation activity. For instance, we have observed this frequently in our experience with @note, where several experts in literature, working on the same text with different (even opposing) purposes, defined different annotation ontologies, and thus different annotation activities. Once each activity was finished, the use of FCA allowed experts to assess to what extent the different ontologies accomplished aspects like adequacy to the intended annotation tasks, understandability and usability by annotators, etc. The conclusions obtained were very valuable for subsequent discussion among experts and for comparison of different (even divergent) approaches to the organization of digital resources.

## 2.4 Concept lattices for collections of resources annotated according to taxonomical ontologies

FCA can be straightforwardly applied to our approach by assigning digital resources as the objects and ontology concepts as the attributes. It is important not to confuse ontology concepts, which are one of our primary sources of information, with the formal concepts that are the final entities obtained by combining the digital resources and the ontology concepts via FCA. In this model, *ontology concepts* become *attributes* of the formal contexts.

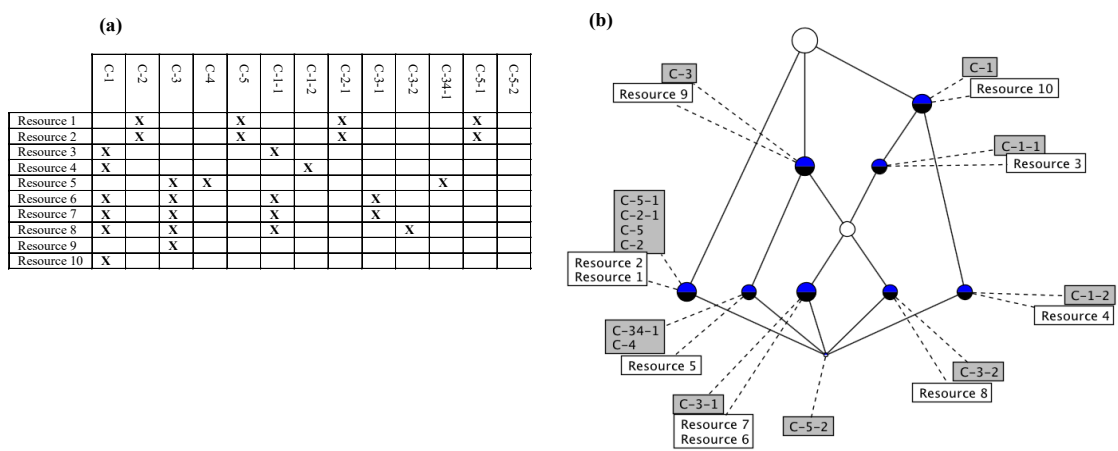

Figure 4. (a) Formal context derived from the ontology in Figure 3; (b) the concept lattice associated with the context in (b)

As we consider our ontologies as taxonomies, we must also include this ontological assumption and define the formal context accordingly. For example, if an annotation is tagged with an ontology concept $OC$, it will also be associated with any other ontology concept $OC'$ such that $OC \subseteq OC'$. With these assumptions, concept lattices are constructed as follows:

- The upper part of the lattice is formed by the attribute concepts associated with each ontology concept $OC$ in the original ontology. Thus, it is easy to recognize the source ontology in the upper segment of the emerging concept lattice with the exception of the ontology concepts that: a) were never used in the annotation process (i.e., they will appear in the bottom section of the lattice); b) always co-occur in the same annotations (i.e., they share the same attribute concepts); and c) although not related in the original ontology, are used together in some cases by the annotators. This structure facilitates the discovery of new relations not present in the original ontology via the final concept lattice.

- The bottom part of the lattice is associated with the formal concepts that are a combination of the upper attribute concepts related to primary ontology concepts. The latter concepts are very interesting, as they reflect how the annotators have combined the original ontology concepts.

To illustrate this process, Figure 4a shows a formal context depicting a possible way in which the annotators used the ontology displayed in Figure 3. The annotated resources are the formal objects and the ontology concepts are the formal attributes. At a glance, the formal context shows the ontology concepts shared by more than one annotation. The formal context maps not only the ontology concept used in the annotation but also the more generic ontology concepts that are super-concepts of the ones selected. For example, if the annotator used the ontology concept C-1-1 to annotate, the formal context will also show its most generic ontology concept as a relation (i.e., C-1). Figure 4b shows the concept lattice and the 11 formal concepts obtained from the formal context. This lattice shows the following:

- The upper part maps the generic ontology concepts that have not been combined with any other ontology concept in any of the annotations. This means that, in those cases, the annotator applied the original ontology. This is the case of the attribute concepts C-1 and C-3. It can be read from the lattice that these ontology concepts have been used in isolation (i.e., for resources 9 and 10) or co-occurring with their ontology sub-concepts (i.e., for resources 3 and 4).

- Ontology concepts C-2 and C-5 always co-occur in the same annotations. In this situation, both ontology concepts will be merged into the same attribute concept: ({*C-2*, *C-5*, *C-2-1*, *C-5-1*}, {*Resource* 1, *Resource* 2}), suggesting that they have a close conceptual relation from the annotator's point of view. Thus, on the basics of the quantitative information, or by examining the actual annotations, domain experts could decide, e.g., to change the original ontology to reflect this situation, thus melding C-2 and C-5 into a single concept, or otherwise to instruct annotators to better clarify the distinction between C-2 and C-5.

- The formal concept {(*C-3*, *C-4*, *C-34-1*), (*Resource* 5)} correctly maps the nature of the ontology concept *C-34-1* with the two aforementioned parents.

- The bottom concept shows the ontology concepts not used by the annotators, the ontology concept *C-5-2* in this case. Domain experts can use this evidence to modify the ontology or to instruct annotators.

- The remainder of the formal concepts depicts situations where the annotators have combined conceptually different ontology elements. For example, ontology concept *C-3-1* has been used in combination with ontology concept *C-1-1* (for resources 6 and 7), as well as ontology concept *C-3-2* (for resource 8). Eventually, domain experts could consider examining the lattice and the associated annotated resources to potentially refine the ontology by assigning suitable names to these emerging conceptual combinations.

All these considerations are examples of the iterative approach to the formulation of annotation ontologies derived from our approach: using FCA, experts are able to assess how ontologies were actually used by annotators; the resulting analysis helps perform structural refinements on these ontologies.

## 3    Assessment of Annotation Activities in @note

This section shows how the approach described in this paper has been implemented in @note, an application for the collaborative annotation of digitized literary works. In subsection 3.1, we summarize the @note application. In subsection 3.2, we describe the assessment of the @note annotation activities using FCA.

### 3.1    The @note application

The application @note is an RIA for the collaborative annotation of digitized literary texts (Gayoso et al. 2012, 2013). This collaborative annotation tool adds free-text notes to literary works and classifies these notes with concepts selected from ontologies specifically devised for the annotation of the specific works. As a result, @note enables the collaborative creation of specialized knowledge bases of notes added to a literary work for a given purpose (e.g., to enable critical reading).

The keystone concept in @note is the *annotation activity*. An annotation activity is primarily characterized by the literary work to be annotated, and the annotation ontology to be used during the annotation of the work. These activities are defined by experts in literature (teachers, researchers, etc.), who can collaborate in the selection of the volume to be annotated and, more importantly, in the definition of the annotation ontology. In addition, these activities are targeted to groups of annotators (students, other scholars, etc.) responsible for adding and cataloguing notes.

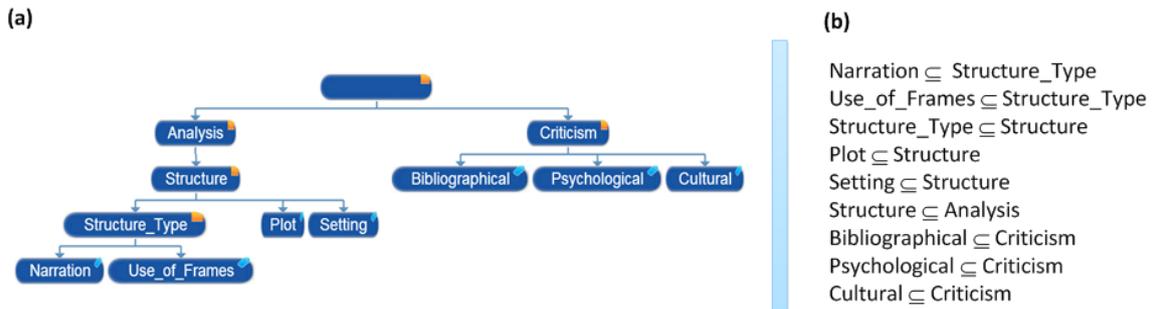

Figure 5. (a) Sample annotation ontology in @note and (b) description logic counterpart of (a).

Following the assumptions used in this study, annotation ontologies in @note are hierarchical arrangements of concepts organized according to an *is-a* relationship. The @note application organizes these concepts into two different groups as follows:

- *Annotation types*, which are the more specific concepts (the leaves in the hierarchy), and are actually those concepts that can be used for classifying notes; and

- *Annotation categories*, which are more general concepts (inner nodes in the hierarchy) that can be specialized in other, simpler categories and/or annotation types. These concepts are used solely for structuring purposes. (@note does not allow them to be used directly for semantically describing notes, although they can be used for searching and browsing.)

Figure 5a shows an excerpt of an annotation ontology in @note, and Figure 5b shows its description logic counterpart.

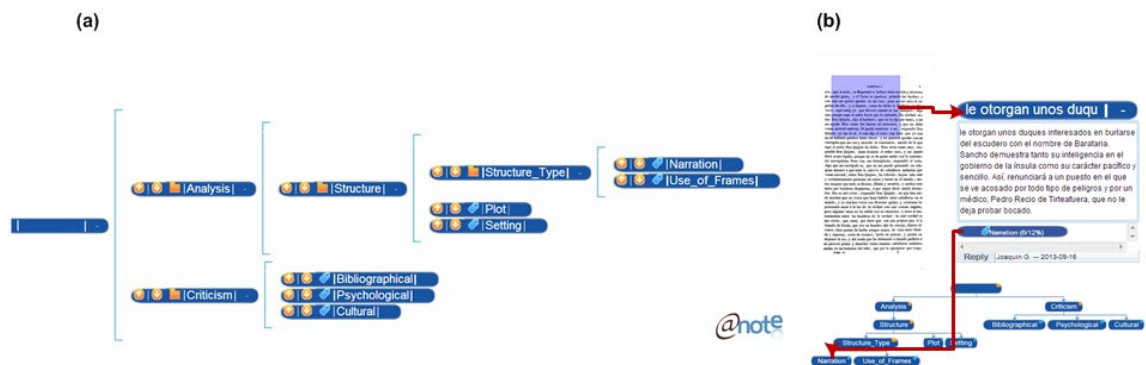

Figure 6. (a) The ontology editor in @note and (b) the annotation of free-text notes in @note.

The ontological bias in @note has been shown to be sufficiently useful to experts and annotators in literature and simple enough to facilitate the collaborative authoring of annotation ontologies by experts. The experts can share their ontologies with other experts and directly edit these ontologies by using a simple tree-based editor integrated in the RIA (Figure 6a). Annotators can also interact easily with the annotation ontologies by using a graph-like view (Figure 6b); see the works of Gayoso et al. (2012, 2013) for more details on the @note functionalities.

### 3.2 *Using FCA to assess annotation activities in @note*

To support the assessment approach described in this paper, @note associates a formal context with each annotation activity as follows:

- The set of objects is composed of the set of all the free-text notes added to the literary work (i.e., the digital resources in this case are composed of the digital notes added to the literary works), and

- Following the directives described in section 2, the attributes associated with each note are composed of all the annotation types that tag the note, as well as all the annotation categories in which these types are included.

Experts can further project this formal context on subgroups of annotators, and even on the notes authored by individual annotators, to better assess the use of the ontology by the subgroups and individual users. Thus, by applying FCA in these formal contexts, @note is able to display how the annotation activity was performed by using Hasse diagrams (see Figure 7a). Each node in the diagram displays the set of new ontology concepts in the annotation ontologies referred to by the node. As shown in Figure 7a, such a set can be empty (as it is in the case of newly-formed concepts not present in the original ontology). Each node can be expanded to show complete information about the formal concept (see Figure 7b) as follows:

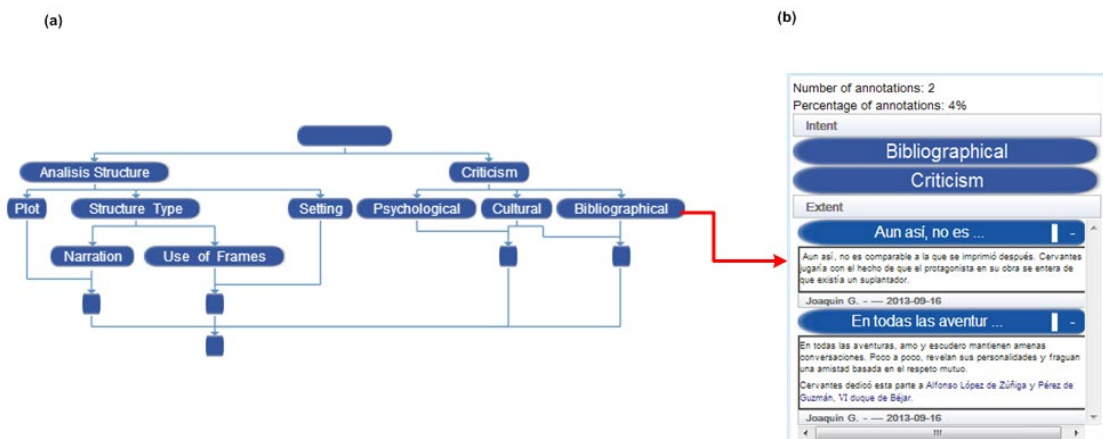

Figure 7. (a) A Hasse diagram depicted in @note and (b) the expansion of a node from (a).

- Quantitative information on the cardinality of the formal concept's extent (i.e., count and percentage of notes included in the concept),

- The formal concept's intent (i.e., annotation types and induced annotation categories), and

- The concept's extent (i.e., the notes grouped in the concept).

This tool thus directly supports the assessment of annotation activities in @note by applying the considerations outlined in section 2.

## 4    Case Study: The Annotation of "The Library of Babel"

This section describes how the FCA-based assessment of annotation activities of @note is applied in practice. For this purpose, we focus on the annotation of *The Library of Babel*, a short story written by the Argentinian writer Jorge Luis Borges (Borges 1944).

### 4.1    The annotation activity

To provide the annotation ontology for this activity, domain experts (teachers, in this case) applied *close reading* as the basic methodology for the critical literary text analysis (Lentricchia & Dubois, 2003). *Close reading* achieves text interpretation through its primary formal and thematic aspects. *Close reading* promotes several readings of the text: *simple reading* (the reader is able to follow the course of the story), *detailed reading* (the reader achieves an in-depth understanding of the text by consulting, e.g., external references, to clarify some text content or to provide additional information), and *interpretative reading* (the reader is able to interpret the contents of the story and to derive ontological and philosophical consequences, e.g., in the case of Borges' story, posing philosophical conjectures about human existence).

Domain experts provided the ontology shown in Figure 8a by applying *close reading*. For this purpose, they first analyzed how *close reading* could be particularized in this text, which led them to formulate a repertory of concepts oriented to the different dimensions of critical text analysis theory. Then they organized these concepts in a meaningfully conceptual hierarchy. This task was facilitated by the collaborative ontology edition mechanisms included in @note (basically, collective edition of the ontology, and use of discussion forums to resolve disagreements on how to materialize *close reading* for the Borges' work).

Concerning the anatomy of the ontology, formal aspects of the text are captured by annotation types in the *References* annotation category. Annotation types include *Authors* referred to in the text, *Citations* made in the text, *Books* referenced, and *Word Meanings* for significant words. Thematic aspects can be catalogued with annotation types under *Characters*, *Time*, *Space* and *Morals*, as well as with the *Authorities* annotation type. Thus, the goal of the domain experts developing this ontology is to allow the annotators (students, in this case) to see how, according to Borges' literary imagination, the *Library* (*Space*), where all the *Books* in the world (and therefore all the *Authors* and all the cultures, represented by *Citations* and *Word Meanings,* anywhere in *Time*) form the *Universe*. Borges' *Universe* includes both a microcosm, the human being, and a macrocosm, divinity. Thus, the order resulting from the union of *Time* and *Space* induces *Morals* (represented by the concepts of *God* and *Devil*). This type of order is perceived only from the *Narrator's* point of view, since the *Other* characters in the story depend upon the *Narrator's* own point of view.

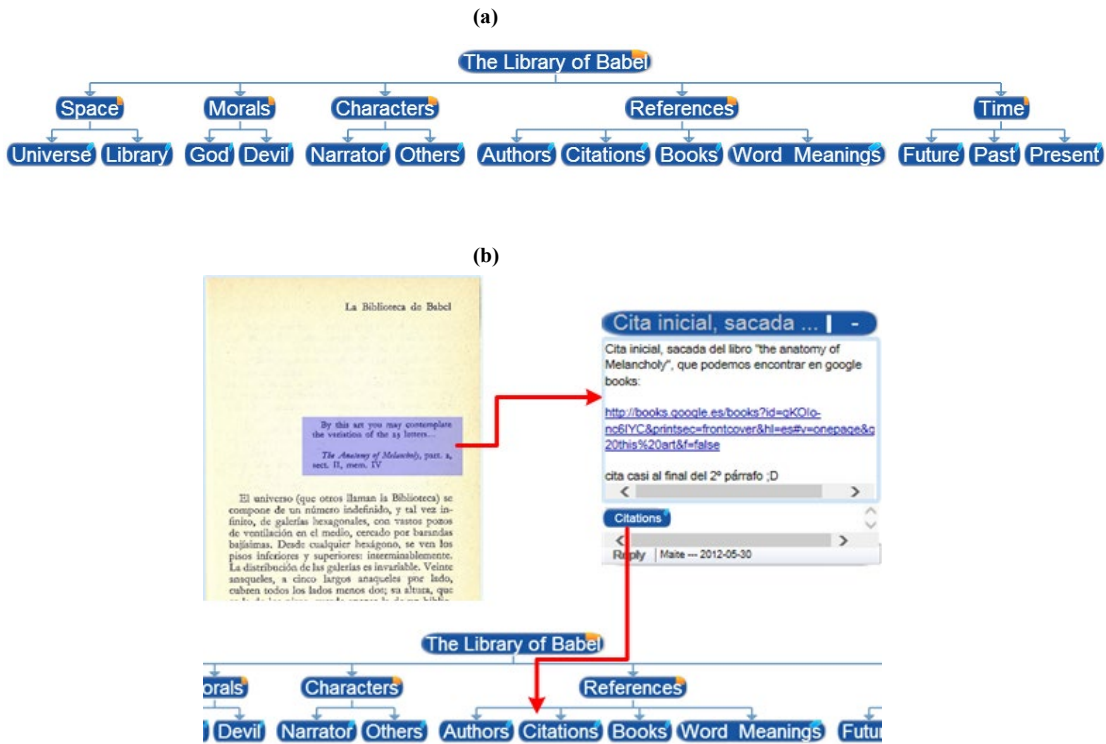

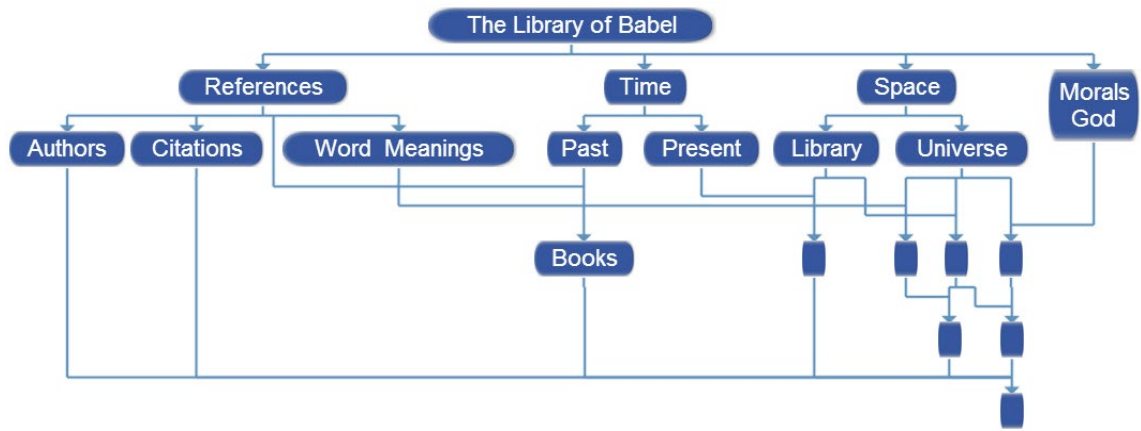

Figure 8. (a) The annotation ontology for "The Library of Babel" and (b) an example of an annotation.

During this annotation activity, annotators (students in this case) created 75 free-text notes, and they catalogued the notes using the ontology provided by the domain experts (their teachers, in this case). Figure 8b shows an example of a note from the ontology catalogued by the students with the annotation type *Citations*.

Figure 9. The concept lattice for the annotation of the *Library of Babel*.

### 4.2 Assessment of the activity

When the annotation activity was complete, annotation experts assessed it by using FCA. Figure 9 shows the concept lattice obtained by applying FCA to the annotation activity depicted in @note. The figure shows that the situations described in section 2.3 are also present in this scenario. The following examples illustrate the different scenarios:

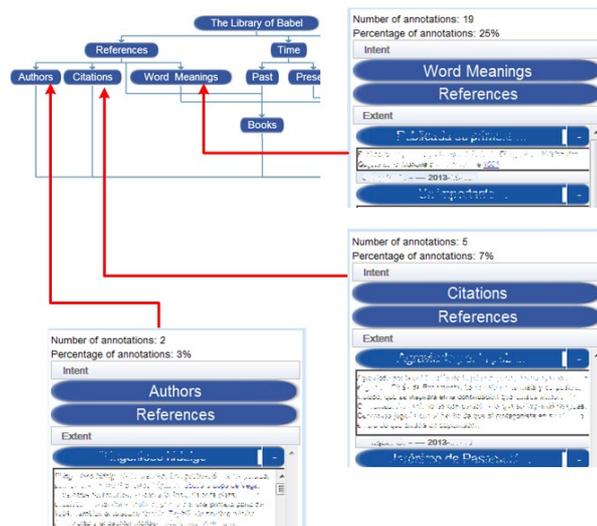

Figure 10. The formal concepts involving the *Authors*, *Citations* and *Word Meanings* ontology concepts.

- The annotation types *Authors*, *Citations* and *Word Meanings*, which are all in the *References* annotation category, were used in isolation (they were used in 3%, 7% and 25% of the notes, respectively) (Figure 10). This is evidence that the ontology was used at the *detailed reading* level of the *close reading* method, i.e., the annotators found unknown authors, citations and words in the text, then they consulted external sources on the Internet (e.g., web pages), then they created notes explaining these elements, and they classified these notes in one of these concepts (depending on whether the unknown term was an author, a citation, or another word).

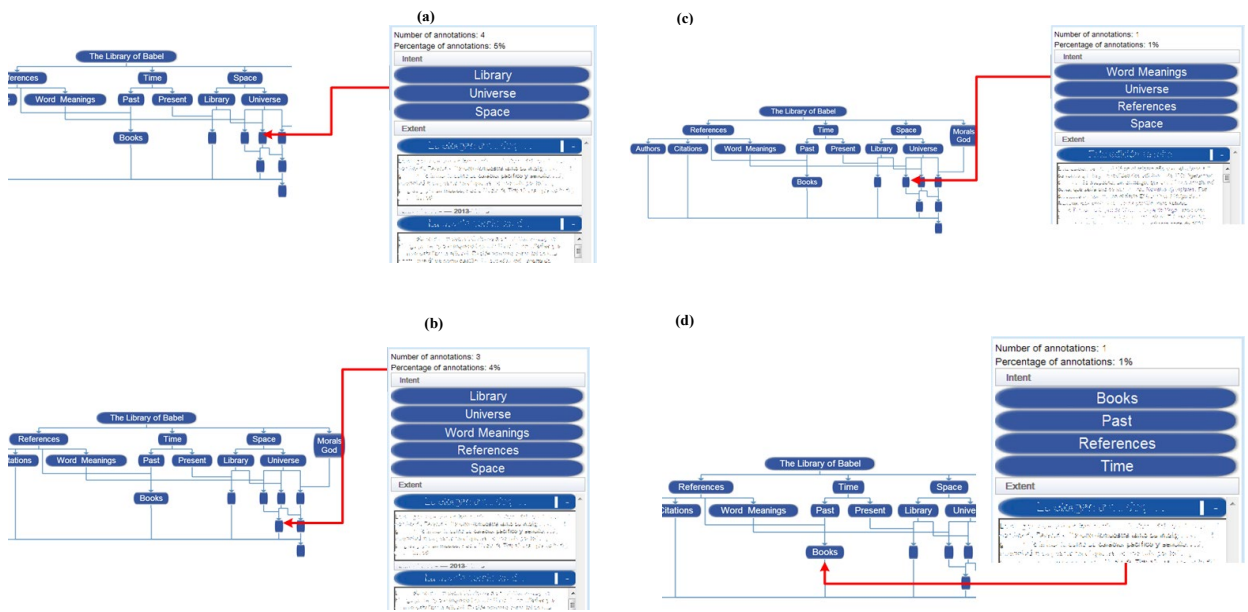

Figure 11. (a) The combination of *Library* and *Universe*, (b) the combination of *Word Meanings*, *Library* and *Universe*, (c) the combination of *Word Meanings* and *Universe*, and (d) the combination of *Books* and *Past*.

- The combination of concepts is evidence that the ontology was used at the *interpretation level* of the *close reading* method. The most prominent example of this is the combination of *Library* and *Universe* in 5% of the notes (Figure 11a). It is a straightforward ontological interpretation of Borges' text, where the library and the universe share the same time/space/morals paradigm. The combination of *Word Meanings*, *Library* and *Universe* in 4% of the notes reveals a deeper interpretation, in which the term *infinite* is explained, not in terms of its objective (dictionary-based) definition, as would be done at the *detailed reading* level, but in terms of Borges' association of *Library* and *Universe* (Borges' *Library* is infinite, and so is the *Universe*, thus *Library* and *Universe* can be identified) (Figure 11b). The combination of *Word Meanings* and *Universe* in 1% of the notes follows a simpler interpretation of the association, in which *Library* has been omitted because of the prior *Library-Universe* identification (Figure 11c). After examining the notes' contents, the combination of *Books* with *Past* in 1% of the notes revealed (Figure 11d) a philosophical interpretation. Indeed, the pilgrimage of the narrator during his youth, i.e., the *Past*, in search of the *Book,* was interpreted as a metaphor for the meaning of life, wherein life is circular, cyclical and infinite, as are the universe and the library in Borges' literary imagination.

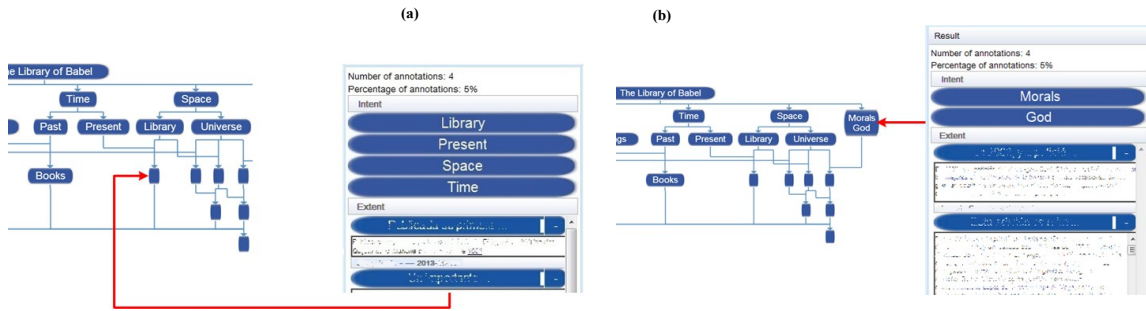

Figure 12. (a) The combination of *Library* and *Present* and (b) the formal concept for *God*.

- Other combinations of concepts resulted in incorrect interpretations of the text, however. For example, 4% of the notes were tagged with *Library* and *Present* (Figure 12a). After examining the notes, experts realized that these notes were from the same annotator, who incorrectly related the infiniteness of the Borges' Library with the present, interpretations that could hardly be derived from Borges' text (rather, the library is infinite, and therefore timeless).

- The experts also examined the unused concepts in the activity (those downgraded to the bottom concept), i.e., *Future*, *Narrator*, *Other*, *Devil* and *Authorities*. With respect to *Future*, its lack of use is explained in this particular text because time in the Borges story oscillates between past and present, although the concept could be useful for other works. Regarding the lack of notes involving aspects of the characters, experts realized that to focus this category in terms of particular characters might be unproductive because it is too generic a term (in particular because Borges' text is narrated in the first person, characters other than the narrator are subjected to the narrator's point of view). Similarly, experts recognized the sub-categorization of *Morals* into *God* and *Devil* could lead to a misconception. Indeed, the goal was not to recognize elements concerning deity; although in the text, it is possible to make some interpretations concerning *God* (5% of the notes did, as shown in Figure 12b), it has nothing to do with *Morals*. With respect to *Devil,* the text does not contain any mention of the *Devil*, only *Evil*. The experts also realized that the *Authorities* annotation type is meaningless for this type of work because it can be superseded by using other annotation types in *References*.

### 4.3    Corrective Actions

After completing the assessment, the domain experts performed the following corrective actions based on the results of the FCA applied to the annotation activity:

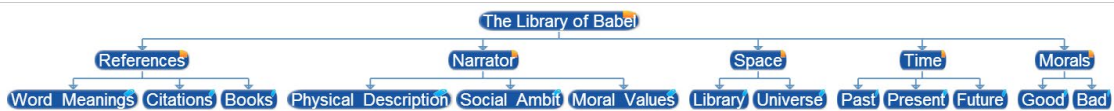

Figure 13. The evolution of the annotation ontology as a result of the assessment of the annotation activity.

- The domain experts warned annotators about the potential misconceptions. In particular, authors of the notes in Figure 11c were notified of the possibility of including *Library*, in addition to *Universe* and *Word Meanings*, when discussing the role of *infinite* in the text. They also warned the author of the notes associated with Figure 12a about the incorrect combination of *Library* and *Present*.

- They also refined the annotation ontology. Rather than providing concepts for the different types of characters (*Narrator* and *Other*), they decided to focus the ontology on the analysis of the first-person text, therefore providing concepts for tagging the narrator's different features, i.e., *Physical Description*, *Social Ambit* and *Moral Values*. They decided to eliminate *Authorities*. They chose more abstract names for the *Morals* concepts, i.e., *Good* and *Bad* instead of the more anthropomorphic ones (*God* and *Devil*). The resulting ontology, created in @note, is shown in Figure 13.

## 5 Method Evaluation

To evaluate the usefulness of the method, we executed an informal survey among domain experts who were using this method in @note. In the survey, we focused on the following three essential aspects of the method:

- *Does the method allow domain experts to assess whether annotators are using the ontology appropriately in the context of the annotation activity*? The response to this question was mostly positive. Experts highlighted the importance of the explicit visualization of the combinations between concepts as one of the key aspects in assessing the use of the ontology during annotation. They also mentioned the possibility of projecting concept lattices onto specific subgroups of annotators and even onto individuals, offered by @note as an essential and very useful feature (in particular in educational settings, where @note is currently used). As a potential feature to be added to the tool, they suggested the inclusion of semi-automatic assessment support (e.g., by adding rules able to detect good and bad uses in the ontology) and the inclusion of dynamic/continuous assessment, so that intermediate assessment results can impact the annotation activity. They also emphasized the specific nature of the @note domain (annotation of literary texts) and the specific purpose of the activities in @note (empowering the reading of literary works through explicit annotation activities), warning us that what is valid for this domain may not necessarily be extrapolated to other annotation domains. Nevertheless, they also agreed that, given the complexity of this domain and of the intended activity (human reading), the possibility of extrapolation is more than a reasonable assumption.

- *Does the method allow domain experts to help annotators make better use of annotation ontologies*? While the experts agreed that the method provides valuable information to help annotators use ontologies more effectively (as the case study in the previous section illustrated), they also agreed that it should be accompanied by better support for offering feedback to the annotators (e.g., informative messages linked to ontology concepts explaining their intended use in the annotation activity). They also noted the need to personalize feedback based on the annotators' expected expertise (this feature is very important in educational settings). The inclusion of mechanisms to allow/forbid certain combinations (e.g., using rules) would also be a welcome improvement.

- *Does the method allow domain experts to enhance their ontologies after the annotation activity*? The response to this question was unanimously affirmative. After applying the method, domain experts refined their annotation ontologies in similar ways to those described in the case study (erasing useless concepts, rethinking concept sub-hierarchies, choosing better names for concepts, etc.). The case study is a clear example of this positive outcome.

## 6 Related work

Most of the research on the use of artificial intelligence and knowledge-based techniques for enhancing the annotation activities of online resources has been focused on the automation of the semantic annotation process (i.e., the automatic addition of semantic annotations to digital resources), in the context of the semantic web, and in the annotation of text resources (e.g., text documents and HTML pages) (Reeve & Han, 2005) (Oliveria & Rocha, 2013). Typically, these systems use natural language processing techniques to identify the parts of the documents to be annotated and, ideally, to determine the content of the annotations. To acquire and exploit the additional linguistic knowledge required for the annotation process, these systems can adopt different strategies, as described below:

- *Data-driven* strategies. In these strategies, the annotation process is orchestrated by using solely the corpus of pre-annotated texts. An example of a system following this strategy is SemTag, the semantic annotation component of the Seeker platform (Dill, 2003; Dill et al., 2003), which tags parts of the documents with concepts taken from a taxonomy. The only additional knowledge required by SemTag is the annotated corpus to determine the corpus-wide distribution of terms at each node of the taxonomy.

- *Knowledge base* strategies. According to these strategies, the user manually provides knowledge bases that contain the additional knowledge required to perform the annotation. These knowledge bases can conform to formalisms such as regular expression-based patterns or rules. Examples of systems following this strategy are AeroDAML (Kogut et al. 2001), a system used to annotate documents with the DAML agent annotation language (Greaves, 2004), KIM (Popov et al. 2003; Popov et al. 2004; Malik et al. 2010), a flexible platform for pattern-based semantic text annotation, MUSE (Maynard, 2003), a system for named entity recognition and co-referencing in text documents, Cerno (Kiyavitskaya et al. 2007; Kiyavitskaya et al. 2009), a system for domain-specific text annotation, KnowWE (Baumeister et al. 2008; Baumeister et al. 2010), a semantic wiki system, Lixto (Baumgartner et al. 2007), which helps users in the creation of rule-based wrappers and which uses a visual style, and Semantic Wikipedia (Krötzsch et al. 2006), a framework for extracting semantic information from plain text.

- *Pattern-discovering* strategies. According to these strategies, the user provides an initial knowledge base made of patterns, which is automatically extended (Brin, 1999). Armadillo (Kiyavitskaya et al. 2009, Dingli et al. 2003), a system for the domain-specific annotation of large repositories of texts, or PANKOW (Cimiano et al. 2004a), a pattern-based annotation algorithm, are practical examples of these strategies.

- *Wrapper induction* strategies. These systems use machine-learning techniques to induce wrappers to locate the sections in the documents to be annotated (Kushmeric, 2000). Usually (although not necessarily) these wrappers are described as automatically-induced knowledge bases. Examples of systems using wrapper induction are MnM (Vargas-Vera et al. 2002, Vargas-Vera et al. 2007), a semantic annotation environment able to infer tagging rules from an annotated corpus, Ont-O-Mat (Handschuh et al. 2002), a system implementing the CREAtion of Metadata (CREAM) framework (Handschuh et al. 2002, Handschuh et al. 2002), GoNTogle (Giannopoulos et al. 2010), a system that exploits both the text structure and previous annotations made by users, KnowItAll (Etzioni et al. 2005), a system oriented to extracting large collections of facts from the web, and Thresher (Hogue et al. 2005; Huynh et al. 2002), a system able to produce wrappers from examples provided by the end user.

Contrary to the approach used in the above systems, the use of artificial intelligence techniques in the method described in this paper is oriented to analyzing the resources annotated, not to automating the annotation process. It is especially well suited to domains where annotations cannot easily be inferred from the contents automatically (e.g., annotation of literary texts according to some literary criteria, such as the case study presented in this paper illustrates). It is also especially well suited to annotation activities comprising non-text data, such as the annotation of images, video and other media (Dasiopoulou et al. 2011), or the annotation of text resources using free-text notes with a clearly defined scholarly purpose (e.g., critical text analysis), as enabled by tools similar to @note (Azouaou & Desmoulins, 2006; Koivunen 2005; Rocha et al. 2009; Schroeter et al. 2006; Tazi et al. 2003; Donato et al. 2013). Additionally, because this method is focused on the assessment stage of the annotation process, it could be meaningfully combined with any of the previously mentioned massive annotation methods.

FCA has been used as a primary tool in ontology engineering for different purposes (Ciminano et al., 2004b; Poelmans et al. 2013) as follows:

- *Ontology construction*. This activity addresses the construction of ontologies for a given domain. For this purpose, FCA supports bottom-up approaches to ontology construction, in which the process focuses on building formal contexts with concept instances as objects and features of concept instances as attributes. Xu & Xiao (2009) exemplify this approach in the computer network management domain; Bao et al. (2005) exemplify it in the domain of pressure component design; and Chi et al. (2005) in the domain of digital archives. Richard et al. (2004, 2006) have developed a method to combine rule bases and FCA in the construction of ontologies. In their method, classification rules characterizing objects are used, which makes it possible to identify rules as objects and rule conditions as attributes in the formal concepts. Another typical application of FCA in ontology construction is to build initial ontologies from a set of documents. In this case, documents are preprocessed using standard natural language processing techniques, and then ontologies are built from the result of this preprocessing (Jiang et al. 2003; Soon & Kuhn, 2004; Xu et al. 2006; Gamallo et al. 2007; Bendaoud et al. 2008). Kiu & Lee (2008) use FCA to edit existing ontologies (i.e., to add, delete and modify existing concepts) instead of constructing ontologies from scratch.

- *Ontology enhancement and quality management*. This activity addresses ontology refinement to better suit the target domain. Rudolph (2004) uses FCA to add axioms, in the form of implication rules, incrementally and interactively to a description logic-based representation of an ontology. He focuses on a finite universe of objects and on pairs of these objects. Then, he uses ontology concepts as attributes for these objects and ontology roles as attributes for the pairs. By using the associated concept lattices to approximate hypothetical axioms and by asking domain experts about the validity of these axioms when they are not covered by the current ontology, the method can enrich the ontology with new axioms or otherwise enrich the formal contexts with appropriate counterexamples. Rudolph et al. (2007) and Völker and Rudolph (2008) extended the technique and combined it with natural language processing to cope with the refinement of lexical ontologies, and Rudolph (2008) extended it to acquire complete sets of domain-range restrictions. Sertkaya (2009) used a similar approach to complete ontologies with relevant information about a domain. Kim et al. (2007) considered sets of ontology constructs as objects and sets of binary relations as attributes. They then mapped ontologies onto formal contexts and applied FCA to detect potential problems in the ontologies. Jiang et al. (2009a) and Jiang et al. (2009b) used FCA to audit the quality of two real-world ontologies.

- *Ontology mapping and merging*. Ontology mapping is the transformation of source ontologies into target ones, i.e., with knowledge representations of overlapping fields likely to represent the same concept with different names, while ontology merging is the amalgamation of several ontologies into a single one. These two interrelated activities have also been addressed by using FCA. Stumme & Maedche (2001) proposed a merging method focused on concept instances. These instances were used to construct formal contexts, which, in turn, were merged. The resulting concept lattice was pruned using information from the original ontologies, and, finally, the merged ontology was generated from the pruned lattice with the help of the domain experts. De Souza et al. (2006) solved the interoperability issues of overlapping ontologies by extracting similarity measures for the identification of concepts related across ontologies. They used thesauri as a bridge representation, i.e., they associated terms in thesauri with concepts in ontologies, and then mapped the thesauri in concept lattices. Similarity distances as defined in terms of the resulting lattices. Fan & Xiao's (2007) approach focused on similarity measures between ontologies in terms of subclass mapping, rather than in terms of entity. To do so, it computed inclusion measures to map the ontologies. Other works have used ontologies to uncover similarities between FCA concepts, as in Formica (2006). Zhao et al. (2007) proposed transforming ontologies into formal contexts and then merging them to obtain a concept lattice, while concurrently developing a similarity measure based on a rough FCA. Le Grand et al. (2009) applied FCA to complex systems analysis, and Krotzsch et al. (2005) proposed modeling complex relationships by using morphisms to formalize the interplay between two knowledge bases.

Thus, while most of the works dealing with the use of FCA in ontological engineering are focused on ontology management, assuming that FCA can facilitate ontology management operations such as merging, mapping, assessment, and quality assurance, our approach is more focused on the annotation activities themselves. Domain experts use concept lattices induced by annotated resources and the structural organization of ontological concepts to assess particular annotation activities. In consequence, they can either instruct annotators on the better use of ontologies, enhance annotation ontologies (as many of the aforementioned works on the use of FCA in ontological engineering do), or adopt a mixture of both types of corrective actions.

## 7   Conclusions and future work

The semantic annotation of collections of digital resources enhances the cataloguing and retrieval of the resources and, more importantly, enables a more sophisticated use of these resources in different applications. Semantic annotation can require both standardized annotation schemas and domain-specific ontologies specifically designed by domain experts to suit the features of the collection and the intended use of the resources therein. However, in order to ensure the quality of the annotations it is necessary to assess to what extent annotators made full use of the ontologies, and to what extent the ontologies provided were suitable to the annotation task envisioned. By doing so, domain experts can, on one hand, advise annotators on how to improve their ontology usage. On the other hand, domain experts can detect aspects from the ontology that can be improved in order to better meet annotation requirements. In consequence, they are able to re-structure their ontologies, which leads to an incremental and iterative process of ontology enhancement. This paper has shown how to achieve these features by using FCA.

From a theoretical point of view, the main contribution of this paper is to developing a generic approach to the assessment of semantic annotation activities, based on FCA. This approach is particularly suited to settings where the annotation of resources cannot easily be automated on the basis of the resource structure, and therefore, must be performed by a community of annotators though a collaborative and iterative process. Since in this approach the responsibility of ontology design is assigned to domain experts, we constrain the ontologies to hierarchical arrangements of concepts, i.e., concepts related by an *is-a* relationship. Other kinds of relationships are intentionally excluded in order to facilitate the authoring of ontologies by using suitable hierarchy editors. In this way, by considering annotated resources as objects and ontology concepts as attributes of a formal context, FCA is used to create a concept lattice from annotated collections of digital resources. The upper part of the lattice contains the ontology concepts from the original ontology used to annotate the resources, as well as the *is-a* relationships among these concepts, re-structured according to evidence of use gathered from the formal context. The lower part contains the different combinations of ontology concepts meaningfully and distinctively used during the annotation. Thus, by inspecting the lattice, experts can gain insight on how annotators actually used the ontology, uncovering those uses caused by a misconception of the annotation guidelines, and those due to potential problems in the original ontology. In consequence, the aforementioned assessment goals (i.e., to instruct annotators in the better use of the annotation ontology, and to enhance the ontology according to its practical usage) can be achieved.

From a practical point of view, the main contribution of the paper is showing how the proposed approach can be implemented in practice. For this purpose, we have described how the approach has been implemented in @note, a tool for the collaborative annotation of digitized literary works. In addition, we have illustrated how this implementation works in practice, with an annotation activity focused on *The Library of Babel*, a short story written by the Argentinian writer Jorge Luis Borges. In this setting, and in order to evaluate the approach, we ran an informal survey among experts in literature who used @note. The outcomes were mostly positive: domain experts considered the approach valuable to helping annotators improve their annotation skills with respect to established annotation ontologies, and a valuable tool for enhancing annotation ontologies themselves. They also suggested some improvements to the approach, concerning support for greater automation of the assessment process by using rules operating on the concept lattice.

In this way, as main strengths of the approach we can highlight its feasibility and ease of use. Indeed, as the experience of @note with experts in literature has made apparent, domain experts (literature teachers) are able to carry out basic conceptualizations of annotation ontologies in terms of *is-a* arrangements of concepts, they are able to give meaningful interpretations to the resulting concept lattices once annotation activities have finished, and, more importantly, they are able to instruct annotators (students, in this setting) regarding their misconceptions in using the ontology, and to enhance the ontology itself as the result of usage experiences.

Finally, we are aware of some weaknesses in the approach. Perhaps the most significant one is the rather strong ontological assumption made, which confines the ontologies allowed to taxonomical arrangements of atomic concepts. However, as argued earlier, this assumption is necessary in order to maintain the feasibility of the approach concerning the role of domain experts as ontology designers. Whether this assumption can be relaxed without compromising usability may be the object of future inquiries. Another weakness of the approach is whether the visual representation of the lattice scales well for larger ontologies. In this respect, works like that of Katifori et al. (2007) suggest that it is possible to use sophisticated visualization techniques to cope with huge hierarchical structures. However, whether these techniques will be well received by domain experts (and, in particular, by literature experts in the context of @note) deserves more research efforts. Lastly, another weakness in the approach is the limited support for helping domain experts analyze the concept lattice. As indicated above, our experiences with domain experts suggested some interesting directions (e.g., to use rules for automating some aspects of the assessment). However, these aspects deserve further investigation.

Currently we are working on the human-computer interaction aspects of the approach in the context of @note on the basis of domain expert feedback. On the basis of this feedback, we are also adding support for automating some assessment aspects by enabling the definition and use of assessment rules able to detect common situations that demand domain expert attention. For future work, we will address other aspects raised by domain experts, i.e., dynamic/continuous assessment, support for attaching feedback to the annotation ontology, etc. In addition, we are also planning to apply the approach to other settings (repositories of learning objects in the educational domain and a collection of digitized and digital objects in the Digital Humanities scenario). Finally, we plan to work on the weaknesses of the approach mentioned above, improving visualization support for concept lattices, and more in-depth research oriented to relaxing the basic ontological assumption adopted.

## 8    Acknowledgements

This work was funded by Google (Digital Humanities Award Programs 2010, 2011), as well as by the project grants FFI2012-34666, TIN2010-21288-C02-01, TIN2009-14317-C03-03 and S2009/TIC-1650.

We would also like to thank César Ruiz for his work in the development of @note, as well as the members of ILSA and LEETHI research groups for their effort in conceiving and evaluating the application.